\documentclass[final,3p,times,onecolumn]{elsarticle}
\usepackage{graphicx}
\usepackage{xcolor}
\usepackage{listings}
\usepackage{caption}
\usepackage{amsmath}
\usepackage{amsthm}
\usepackage{subcaption}
\usepackage{esvect}
\usepackage{booktabs}
\usepackage{amsfonts}
\usepackage{url}
\usepackage[ruled,vlined,linesnumbered]{algorithm2e}
\begin{document}

\begin{frontmatter}
\title{Explainable Artificial Intelligence for Human Decision-Support System in Medical Domain}

\author[Aalto,Umea]{Samanta Knapič}
\ead{samanta.knapic@aalto.fi }
\author[Aalto]{Avleen Malhi}
\ead{avleen.malhi@aalto.fi}
\author[Aalto,KTH]{Rohit Saluja }
\ead{rohit.saluja@aalto.fi}
\author[Aalto,Umea]{Kary Fr\"amling}
\ead{kary.framling@cs.umu.se}

\address[Aalto]{Department of Computer Science, Aalto University, Konemiehentie 2, 02150 Espoo, Finland}
\address[Umea]{Department of Computing Science, Ume\aa  University, 90187 Ume\aa, Sweden}
\address[KTH]{Department of Electrical Engineering and Computer Science, KTH Royal Institute of Technology, Stockholm, Sweden} 

\begin{abstract}
In the present paper we present the potential of Explainable Artificial Intelligence methods for decision-support in medical image analysis scenarios. With three types of explainable methods applied to the same medical image data set our aim was to improve the comprehensibility of the decisions provided by the Convolutional Neural Network (CNN). The visual explanations were provided on in-vivo gastral images obtained from a Video capsule endoscopy (VCE), with the goal of increasing the health professionals' trust in the black box predictions. We implemented two post-hoc interpretable machine learning methods LIME and SHAP and the alternative explanation approach CIU, centered on the Contextual Value and Utility (CIU). The produced explanations were evaluated using human evaluation. We conducted three user studies based on the explanations provided by LIME, SHAP and CIU. Users from different non-medical backgrounds carried out a series of tests in the web-based survey setting and stated their experience and understanding of the given explanations. Three user groups (n=20, 20, 20) with three distinct forms of explanations were quantitatively analyzed. We have found that, as hypothesized, the CIU explainable method performed better than both LIME and SHAP methods in terms of increasing support for human decision-making as well as being more transparent and thus understandable to users. Additionally, CIU outperformed LIME and SHAP by generating explanations more rapidly. Our findings suggest that there are notable differences in human decision-making between various explanation support settings. In line with that, we present three potential explainable methods that can with future improvements in implementation be generalized on different medical data sets and can provide great decision-support for medical experts.
\end{abstract}

\end{frontmatter}
\section{Introduction}
Traditionally, the possible lesions or signs on the captured images are checked manually by the doctor in the medical setting. This manual approach is time-consuming and relies on the doctor's prolonged attention, who has to look at thousands of images from a single medical procedure. On the other hand, in recent years, in the fields such as  medical diagnostics, finance, forensics, scientific research and education, deep learning and, thus, AI-based extraction of information from images has received growing interest. In these domains, it is a often necessary to clarify the model's decisions in order for the human to validate the decision's outcome  \cite{meske2020transparency}. 

Renectly a number of computer-aided diagnostic (CAD) tools were being developed to allow for the automatic or semiautomated identification of lesions. By automatically extracting feature representation, the newly introduced Convolutional Neural Network (CNN), also known as deep neural network, began producing significantly higher accuracy compared to the standard approaches \cite{coelho2018deep}. The use of reinforcement learning techniques and Deep learning methods trained on massive data sets have surpassed efficiency of the humans, producing impressive results even in the medical field. Through the use of machine learning techniques, the lesion detection method is automated with compelling accuracy saving both time and manual effort \cite{malhi2019explainable}. The well-trained machine learning systems have the potential to make accurate predictions from the individual subjects about various anomalies, and can therefore be used as an effective clinical practice tool. However, even though we can understand their core mathematical concepts, they are considered black-box models which lack an explicit declarative information representation and have trouble producing the underlying explanatory structures \cite{malhi2019explainable}.

As AI becomes more effective and is being used in more sensitive circumstances with significant human implications, trust in such systems is becoming highly essential \cite{meske2020transparency}. Humans must be able to understand, re-enact and manipulate the machine decision-making processes in real time. As a result, there is an increasing need to improve the comprehensibility of machine learning algorithm's decisions that can be replicated for real applications, especially in medicine.  This calls for systems that allow straightforward, understandable and explainable decisions being made, along with re-traceable results on demand. Since medical professionals work with dispersed, heterogeneous, and complex data sources, Explainable Artificial Intelligence (XAI) can help to promote the application of AI and Machine Learning in the medical sector, and in particular, it can help to foster transparency and trust. Hence, before they can be trusted, machine learning models should be able to justify their decisions. The explanation support would help to clarify the decision taken by a black-box model which would be more intuitive for humans. With additional explanation of the decisions made, the method would be more reliable and would thus assist the medical professionals in making the correct diagnosis \cite{malhi2020explainable}.

In the present study, we introduced a neural network, in this case a Convolution Neural Network (CNN),  with a specific application in the field of medicine. With the use of post-hoc interpretable machine learning methods LIME and SHAP and the alternative explanation approach CIU, centered on the Contextual Value and Utility (CIU) method to explain machine learning predictions \cite{anjomshoae2020py}, our aim was to improve the comprehensibility of the decisions provided by the CNN. With the aim of helping health professionals in trusting the black-box predictions, we applied the explanations to in-vivo gastral images that were obtained from Video capsule endoscopy (VCE). The three explanation types applied to the same medical image data set were evaluated using human evaluation. We conducted preliminary human decision-support user studies to determine how well humans can understand the provided explanation support and to examine the effect of explanations on human decision-making. The three user groups were presented with the decision support of the three distinct explainable methods LIME, SHAP and CIU which automatically generate different visually based explanations. The following were our test questions: RQ1: Will Explainable Artificial Intelligence improve the trustworthiness of AI-based Computer Vision systems in the medical domain? RQ2: Can various XAI approaches be used as a human decision-support system, and can their explanation strategies be compared? RQ3) Can users recognize the effectiveness of the generated explanations?

\section{Literature review}
It is possible to consider machine learning models as reliable, however, the effectiveness of these systems is limited by the machine’s current inability to explain their decisions and actions to human users. While there is an increasing body of work on interpretable and transparent machine learning algorithms, the majority of studies are primarily centered on technical users. A recently published paper provides comprehensive survey on Explainable Artificial Intelligence studies \cite{adadi2018peeking}. Guidotti et al. \cite{guidotti2018survey} give a detailed analysis of XAI approaches for describing black-box models, and Anjomshoae et al. \cite{anjomshoae2019explainable} provide a systematic review of the literature on explainable agents.  The foundation of the contextual utility and importance of the features \cite{framling1995extracting}, \cite{framling1996modelisation} is the basis of an early approach to understanding the decision of ML models but with the emergence of Deep learning as a popular data analysis tool, new approaches such as SHAP (SHapley Additive exPlanations ) \cite{shap2019}, LIME (Local Interpretable Model-Agnostic Explanations) \cite{malhi2019explainable}, CIU (Contextual Value and Utility) \cite{framling1996modelisation}, ELI5 \cite{eli5}, VIBI and L2XSkater \cite{skater} have developed to provide explanations for the machine learning models. 

\subsection{Explainable Artificial Intelligence in machine learning}

Xie et al. \cite {xie2020explainable} provide a guide to explainability within the realm of deep learning by discussing the characteristics of its framework and introducing fundamental approaches that lead to explainable deep learning. Samek et al. \cite{samek2017explainable}  researched a deep learning method for image recognition with explainability approach where they propose two methods for explanations of sensitivity to input changes. Choo et al. \cite{choo2018visual} present another insightful perspective of potential directions and emerging problems in explainable deep learning. They discuss implementation possibilities concerning human intepretation and control of the deep learning systems including user-driven generative models, progression of visual analytics, decreased use of the training sets, improved AI robustness, inclusion of external human intelligence and deep learning visual analytics with sophisticated architectures. In the project launched by DARPA \cite{gunning2017explainable}, they provide simple conceptual and example applications on the Explainable Artificial Intelligence's current state of work in the domains of defense, medicine, finance, transportation, military, and legal advice. A machine vision-based deep learning explainable framework was used to investigate plant stress phenotyping in \cite{ghosal2018explainable}, with approximately 25,000 photos using feature maps and unsupervised learning to calculate stress intensity.  Hase and Bansal \cite{hase2020evaluating} presented an example of  human subject tests and studied the impact of algorithmic explanations on human decision making. Their studies are the first to provide precise estimates of how explanations impact simulatability across a broad spectrum of data domains and explanation techniques. They demonstrate that criteria for measuring interpretation methods must be carefully chosen, and that existing methods have considerable space for development. 

\subsection{Explainable Artificial Intelligence in the medical field}

Over the past few years, AI-based image information retrieval has gotten a lot of interest in medical diagnostics. Andreas et al. \cite{holzinger2017we} emphasized the importance of using Explainable AI in the medical field to assist medical practitioners in making decisions that are explainable, transparent and understandable. They predicted that ability to explain the machine learning's decision would support the adoption of machine learning in the medical field. In another article \cite{holzinger2017towards} Holzinger et al. in the context of an application task in digital pathology address the importance of making decisions straightforward and understandable with the use of Explainable Artificial Intelligence. \cite{sahiner2019deep} Sahiner et al. provide an outline of the past and present state of deep learning research in medical imaging and radiation therapy, address challenges and their solutions, and conclude with future directions. Amann et al. \cite{amann2020explainability} investigate the function of XAI in clinical settings and come to the conclusion that, in order to eliminate  challenges to ethical principles, the inclusion of explainability is a necessary requirement. It can help ensure that the patients remain at the center of treatment and can make knowledgeable and independent decisions about their well-being with the help of a medical professionals. A novel explanation technique developed by Ribeiro et al. \cite{lime2016}, Local Interpretable Model-agnostic Explanations (LIME), was proposed as a way to explain the predictions of the classifier in a reliable and interpretable way. The model's versatility is shown by including text and image explanations for a variety of models. In making decisions between models, it supported both experts and non-expert users while evaluating their confidence and improving the untrustworthy models by getting an insight into their predictions. Further in \cite{meske2020transparency}, they address the effect of explainability on trust in AI and Computer Vision systems, through improved understandability and predictability of deep learning based Computer Vision decisions on medical diagnostic data. They also explore how XAI can be used to compare the recognition techniques of two deep learning models Multi-Layer Perceptron and Convolutional Neural Network (CNN).

\section{Background}
Lately, laws and regulations are moving towards requiring transparency from information systems to prevent unintended side effects in decision making. The General Data Protection Regulations (GDPR) of the European Union, in particular, grant users the right to be informed regarding machine-generated decisions \cite{voigt2017eu}. Consequently, individuals who are affected by decisions made by a machine learning-based system may seek to know the reasons for the system’s decision outcome. 

\subsection{Black box predictions}
The main concerns about the machine learning model's decision are if we should trust machine learning models' decisions and in what sense do the machine learning or deep learning models make their decisions. Relevant principles in relation to these questions which refer to the ability to observe the processes that lead to decision making within the model are:
\begin{enumerate}
\item Transparency:  A model is considered transparent if it is understandable on its own which usually implies to the easily interpretable models \cite{roscher2020explainable}. Simpler machine learning models tend to be more transparent and thus inherently more interpretable due to their simple structure, such as models built with linear regression.
\item Interpretability: The ability to describe or provide meaning that is clear to humans is known as interpretability. Models are considered interpretable if they are described in a way that can be further explained, such as through domain awareness \cite{doshi2017towards}. The concept behind interpretability is that the more interpretable a machine learning system is, the easier it is to define cause-effect relationships within the inputs and outputs of the system  \cite{linardatos2021explainable}.
\item Explainability: Explainability is more closely linked to the machine learning system's dynamics and internal logic. While a model is training or making decisions, the more explainable it is, the more human understanding of the internal procedures is achieved \cite{linardatos2021explainable}.
\end{enumerate}
An interpretable model does not imply that humans can comprehend its internal logic or underlying processes; thus, interpretability does not necessarily imply explainability, and vice versa.  Interpretability alone is is not enough and the presence of explainability is also important. To meet these objectives, a new area known as XAI (Explainable Artificial Intelligence) has arisen, with the goal of developing algorithms that are both efficient and explainable.

\subsection{Explainable Artificial Intelligence (XAI)} 
\begin{figure*}[!ht]
\centering
\begin{minipage}{.80\textwidth}
  \centering
  \includegraphics[width=1.0\linewidth]{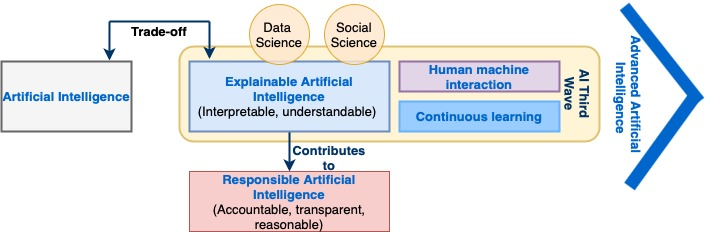}
  \captionof{figure}{XAI basic concepts}
  \label{fig:XAI_basic}
\end{minipage}%
\end{figure*}
Explainable Artificial Intelligence (XAI) methods have been developed in order to achieve greater transparency and produce explanations for AI systems (Figure \ref{fig:XAI_basic}). The Explainable Artificial Intelligence (XAI) research area is investigating various approaches in order to achieve that the behavior of intelligent autonomous systems is understandable to humans. In general, explanations help evaluate the strengths and the limitations of a machine learning model, thereby facilitate trustworthiness and understandability \cite{gunning2017explainable}, \cite{samek2017explainable}, \cite{dovsilovic2018explainable}. One approach of extracting the information of the black-box model’s process of reaching a certain decision are post-hoc explanations. They can provide useful information particularly for practitioners and end-users who are interested in instance-specific explanations rather than the internal working of the model. The goal of XAI models is to create explainable models while maintaining high learning efficiency (prediction accuracy) \cite{meske2020transparency}.  

\subsection{LIME, SHAP and CIU}
Several post-hoc explanation tools for explaining a specific model prediction, including LIME, SHAP, L2X and VIBI have been proposed. One example is the Local Interpretable Model-agnostic Explanations (LIME) explanation tool explaining a model’s prediction by using the most important contributors. It approximates the prediction locally through perturbing the input around the class of interest until it arrives at a linear approximation \cite{lime2016}) and helps the decision-maker in justifying the model’s behaviour. SHapley Additive exPlanations (SHAP) is another example that describes the outcome by “fairly” distributing the prediction value among the features, depending on how each function contributes \cite{lundberg2017unified}. The attributes are as follows: (i) global interpretability – the importance of each indicator that has a positive or negative impact on the target variable;
(ii) local interpretability – SHAP values are determined for each instance, significantly increasing transparency and aiding in explaining case prediction and major decision contributors; (iii) SHAP values can be calculated for any tree-based model \cite{malhi2020explainable}.

Both of the above approaches approximate the local behavior of a black-box system with a linear model. Therefore, they only provide local faithfulness and lose fidelity to the original model. Alternative, the Contextual Importance and Utility (CIU) method for explaining machine learning predictions in principle, states that the importance of a feature depends on the other feature values, in a way that a feature that is important in one context might be irrelevant in another. The feature interaction allows for the provision of high-level explanations where feature combinations are appropriate or features have interdependent effects on the prediction. CIU method consists of two important evaluation methods 1. Contextual Importance (CI) which approximates the overall importance of a feature in the current context, and the 2. Contextual Utility (CU) that provides an estimation of how favorable or not the current feature value is for a given output class \cite{anjomshoae2019explainable}. More specifically the CU gives insight into how much a feature is contributing to the current prediction relative to its importance and alongside the feature importance value the utility value adds depth to explanation.

\section{XAI Methods}

\subsection{ LIME} Our first post-hoc explainability algorithm is Local Interpretable Model-agnostic Explanations (LIME). Ribeiro et al in \cite{lime2016} proposed a local surrogate model LIME  method developed to help users by generating explanations for black-box model's decisions in all instances. LIME's explanation is based on evaluating the classifier model's behavior in the vicinity of the instance to be explained, using the idea of local surrogate models, which can be linear regressions or decision trees, as seen in equation \ref{eqn:Lime}. Here, $x$ is instance being explained. The explanation of $x$ result of maximisation of fidelity term $\mathcal{L}(f,g, \pi_x)$ with complexity of $\Omega(g)$. $f$ represents a black-box model which is explained by explainer represented by $g$. The local surrogate model tries to match the data in the vicinity of the prediction that needs to be explained. Fitting the local model requires enough data around the vicinity of instance being explained, which is done by sampling the data from its neighbourhood.
 
 \begin{equation}
 explanation(x) = argmin_{g \in G}\mathcal{L}(f,g, \pi_x) +  \, \Omega(g)
 \label{eqn:Lime}
\end{equation}

 Initially, LIME uses perturbation technique \cite{lime2016}  to generate samples from original data set. However, in  R \cite{Thomas2019} and Python\cite{Marco2019} implementations, it moved to a different approach. In their implementation, uni-variate distribution of each feature is considered to estimate distributions for each feature where categorical and numerical features are treated differently. For categorical features, sampling is based on probabilities of the frequency of each category. However, for the numerical feature there are three alternatives. First, binning of original data set based on their quantiles and one bin is just randomly picked and sampled uniformly between min and max of the selected bin. Second, LIME approximates the original distribution of numerical features through a normal distribution and the approximated distribution is used to sample the data for that feature. Third, the actual distribution of numerical features is approximated using a kernel density function and sample data from it. LIME utilizes an exponential kernel by design, with the kernel width equaling the square root of the number of features. 

\subsection{SHAP} For the second human evaluation user study, we use Shapley Additive exPlanations (SHAP) as our second post-hoc explainability algorithm to generate the explanations. We looked at Deep SHAP Explainer and SHAP Gradient Explainer, which combine ideas from Integrated Gradients (which require a single reference value to integrate from), SHapley Additive exPlanations (SHAP), and SmoothGrad (which is averaging gradient sensitivity maps for an input image to identify pixels of interest) into an unified expected value equation.We chose the Kernel SHAP algorithm, which is a model-agnostic method for estimating SHAP values for just about any model. We chose the Kernel SHAP algorithm, which is a model-independent method for estimating SHAP values for any model. SHAP KernelExplainer works for all models but is slower than the other model type specific algorithms as it makes no assumptions about the model type. It provided the best results for us, despite being slower than other Explainers and providing an approximation rather than exact SHAP values. The Kernel SHAP algorithm is based on Lundberg et al. paper \cite{lundberg2017unified}  and builds on the author's first paper's open source Shap library \cite{lundberg2017unified} . 

SHAP \cite{lundberg2017unified} aims to explain individual predictions by employing the  game-theoretic Shapley value \cite{shapely1953value}. This approach uses the concept of coalitions in order to compute (as shown in equation \ref{eqn:Shap1}) the Shapley value of features for the prediction of instance ($x$) by the black-box model $(f)$. The  average marginal contribution $(\phi_j^m)$ of feature $(j)$ in all possible coalitions is the Shapley value.  The marginal contribution is calculated as in equation \ref{eqn:Shap2} where $\hat{f}(x_{+j}^m)$ and  $\hat{f}(x_{-j}^m)$ are prediction of black-box $f$ without and  with replace of $j^{th}$ feature of instance $x$ from the sample. 
 \begin{equation}
 \phi_j(x)= \frac{1}{M} \sum_{m=1}^{M} \phi_j^m
 \label{eqn:Shap1}
\end{equation}

\begin{equation}
 \phi_j^m= \hat{f}(x_{+j}^m) - \, \hat{f}(x_{-j}^m) 
 \label{eqn:Shap2}
\end{equation}

\subsection{CIU} For the third human evaluation user study we implement CIU method. Contextual Importance and Utility (CIU) method explains the model’s outcome not only based on the degree of feature importance but also the utility of features (usefulness for the prediction) \cite{framling1996modelisation}. It is consisting of two important evaluation methods: 1. Contextual Importance (CI) which approximates the overall importance of a feature in the current context, and the 2. Contextual Utility (CU) that provides an estimation of how favorable or not the current feature value is for a given output class. 

CIU differs radically from LIME and SHAP because CIU does not create or use an intermediate surrogate model or make linearity assumptions \cite{framling1996modelisation}. Here, Contextual Importance (CI) and Contextual Utility (CU)  are used for generating the explanations and interpretation based on the contributing features of the dataset. It also helps to justify why one class can be preferred over the other. These explanations have \textit{contextual} capabilities, which means that one feature may be critical for making a decision in one situation but irrelevant in another. The mathematical definition (detailed in  \cite{framling2020explainable}) of CI and CU is given in equation \ref{Eq:CI} and equation \ref{Eq:CU} respectively. 
\begin{equation}
CI_{j}(\vv{C},\{i\})=\frac{cmax_{j}(\vv{C},\{i\})-cmin_{j}(\vv{C},\{i\})}{absmax_{j}-absmin_{j} 
} 
\label{Eq:CI}
\end{equation}

\begin{equation}
CU_{j}(\vv{C},\{i\})=\frac{out_{j}(\vv{C})-cmin_{j}(\vv{C},\{i\})}{cmax_{j}(\vv{C},\{i\})-cmin_{j}(\vv{C},\{i\})} 
\label{Eq:CU}
\end{equation}

Here, $CI_{j}(\vv{C},\{i\})$ is the contextual importance of a given set of inputs $\{i\}$ for a particular output $j$ in the context $\vv{C}$. $absmax_{j}$ is the maximal possible value for output $j$ and $absmin_{j}$ is the minimal possible value for output $j$. $cmax_{j}(\vv{C},\{i\})$ is the maximal value of output $j$ observed when modifying the values of inputs $\{i\}$ and retaining the values of the other inputs at those specified by $\vv{C}$. Correspondingly, $cmin_{j}(\vv{C},\{i\})$ is the minimal value of output $j$ observe. Similarly, for contextual utility $CU_{j}(\vv{C},\{i\})$,  $out_{j}(\vv{C})$ is the value of the output $j$ for the context $\vv{C}$.

\section{Methodology}

\begin{figure*}
\centering
\begin{minipage}{.70\textwidth}
  \centering
  \includegraphics[width=0.7\linewidth]{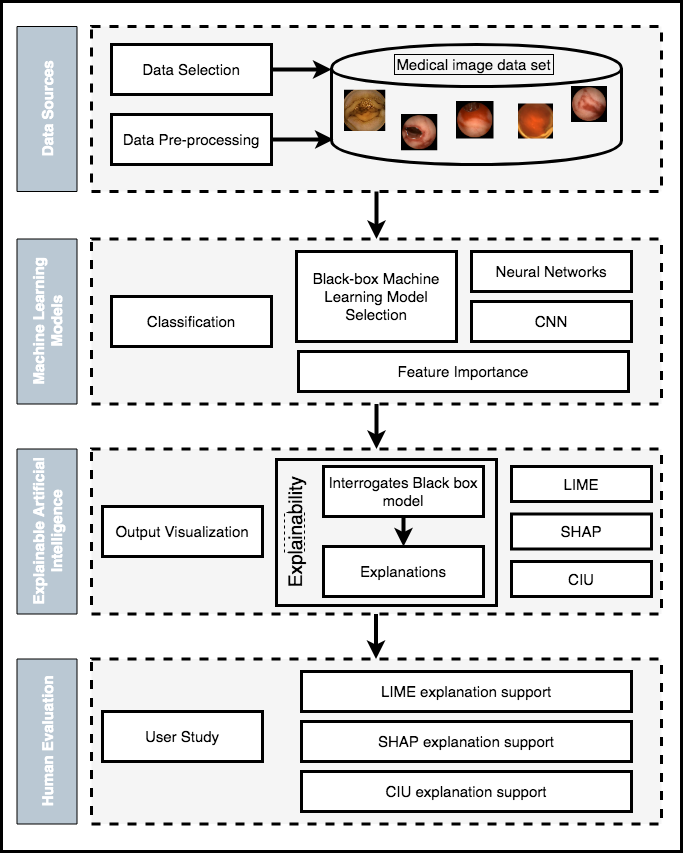}
  \captionof{figure}{Workflow of the proposed method.}
  \label{fig:workflow}
\end{minipage}%
\end{figure*}

This section gives a summary of the methods used to assess the effect of explanations on human decision-making. Figure The entire process is divided into four parts, as shown in Figure \ref{fig:workflow} : data pre-processing, CNN model application, LIME, SHAP, and CIU explanation generation, and assessment of human decision-making in the form of user studies.  Firstly we generate predictions made by machine learning model using the selected data set. In addition, we implement explanations of post-hoc explanation techniques. Methods responsible for assisting human decision-making include Local Interpretable Model-agnostic Explanations (LIME), SHapley Additive exPlanations (SHAP), and Contextual Importance and Utility (CIU). We then evaluate how adding explanations with LIME, SHAP, or CIU affects human decision-making by conducting a user study. 

\subsection{Image data set}
The medical data set considered in the present study is taken from a Video Capsule Endoscopy (VCE), a non-invasive procedure visualizing a patient’s entire gastroenterological tract. The aim of the VCE procedure is to detect segments of red lesion in the small bowel, one of the major organs where bleeding from unknown causes occurs, in order to detect signs of bleeding or polyp. There has been a major breakthrough in diagnosing small bowel diseases with the VCE, however the problem is that there is a ten hours of video material in a single examination which requires a lot of time to read. As a result, analytical methods are needed to improve the diagnosis' efficiency and accuracy. The 3,295 images in the Red Lesion Endoscopy data set were retrieved from  Coelho\footnote{https://rdm.inesctec.pt/dataset/nis-2018-003} \cite{coelho2018deep}. The data set obtains two sets of images. First set contains 1,131 images with lesion and 2,164 without lesion, a total of 3,295 images. The second set contains 439 images with lesion and 151 without lesion, a total of 600 images. Both sets also contain the manually annotated masks marking the bleeding area on each image. All lesion were annotated manually and approved by a trained physician. We focused on Set 1 with a total of 3,295 images, 10\% of which were used for testing and 90\% for training.

\begin{figure*}[!ht]
\centering
\begin{minipage}{.80\textwidth}
  \centering
  \includegraphics[width=1.0\linewidth]{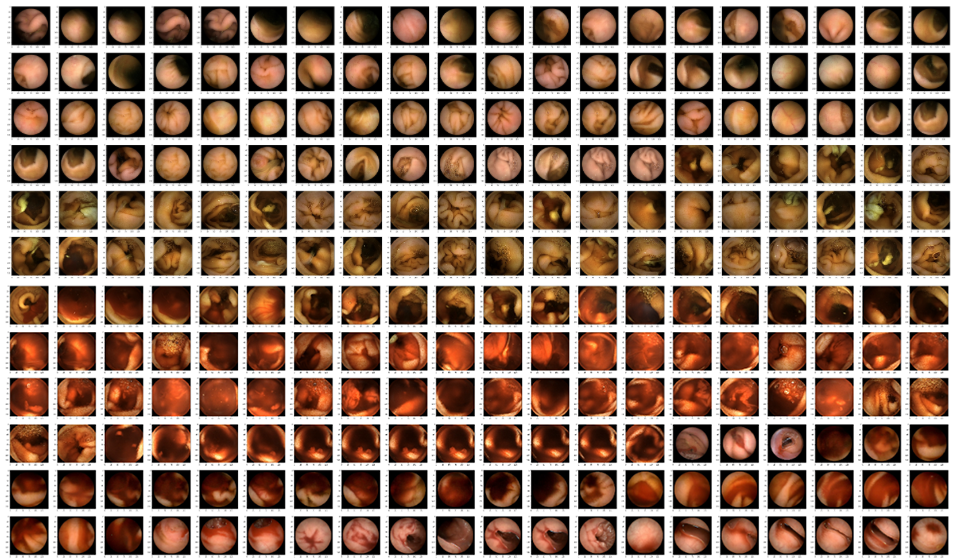}
  \captionof{figure}{Used image data set. Validation part of the bleeding and nonbleeding images.}
  \label{fig:images}
\end{minipage}%
\end{figure*}

\subsection{Implementation of the black box model}

\begin{figure*}[!ht]
\centering
\begin{minipage}{.80\textwidth}
  \centering
  \includegraphics[width=1.0\linewidth]{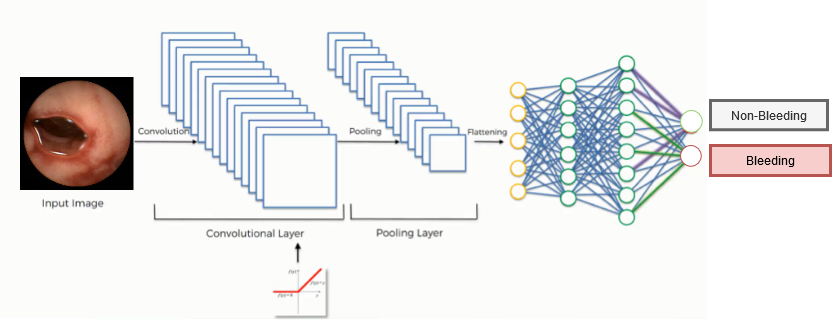}
  \captionof{figure}{CNN model.}
  \label{fig:CNN}
\end{minipage}%
\end{figure*}

\begin{figure*}[!ht]
\centering
\begin{minipage}{.80\textwidth}
  \centering
  \includegraphics[width=1.0\linewidth]{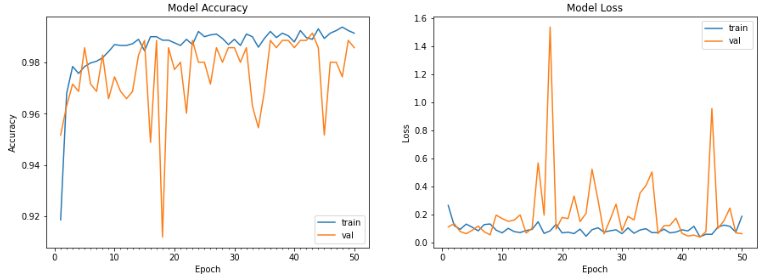}
  \captionof{figure}{Model's accuracy and loss.}
  \label{fig:Model-acc-loss}
\end{minipage}%
\end{figure*}

To begin, we split our data and labels into training and validation sets (randomly assigned). The images are indicative of the medical application situation, as shown in Table \ref{tab:data}, and include both bleeding and non-bleeding (normal) examples. Both 3,295 images have been resized to 150X150 pixels for faster and more accurate computation.  A CNN (Convolutional neural network) model with 50 epochs and a batch size of 16 was used to train the data set (shown in Figure \ref{fig:CNN}) and achieving a validation accuracy of 98.58\%, as shown in Figure \ref{fig:Model-acc-loss}. We trained our CNN model based on labels assigned to each image to recognize the bleeding versus normal (non-bleeding) medical images. The labels were made using the \cite{coelho2018deep} repository's annotated images as a reference point. 
 
\begin{table}[!ht]
\scriptsize
\centering
\caption{Normal (non-bleeding) and bleeding images used for training and testing.}
\begin{tabular}{l|r|r|r}
\hline
Data     & \multicolumn{1}{l|}{\begin{tabular}[c]{@{}l@{}}Normal \\ (Non-bleeding)\end{tabular}} & \multicolumn{1}{l|}{Bleeding} & \multicolumn{1}{l}{Total} \\ \hline
Training & \textit{1940}                                                                         & \textit{1001}                 & \textit{2941}             \\ \hline
Testing  & \textit{224}                                                                          & \textit{130}                  & \textit{354}              \\ \hline
Total    & \textit{2164}                                                                         & \textit{1131}                 & \textit{3295}             \\ \hline
\end{tabular}
\label{tab:data}
\end{table}

The proposed model's architecture is shown in Figure \ref{fig:Model-acc-loss}. We trained our model to recognize the bleeding versus normal (non-bleeding) medical image. The 3,295 images in the data set were split into training and validation sets (randomly assigned). The 3295 data length non-bleeding and bleeding images of 3295 label length were separated into training data of 2941 data length; out of that the bleeding training data length is 1001 and non-bleeding training data length is 1940 and validation data of 354 data length; out of that the bleeding validation data length is 130 and non-bleeding validation data length is 224. The sample images from validation part of the data set are depicted in Figure \ref{fig:images}.
The model provided the output for the medical images assigning them as bleeding or normal (non-bleeding). Table \ref{tab:prob} shows the prediction probabilities for the non-bleeding and bleeding classes that were calculated for a few of the sample non-bleeding and bleeding images. The evaluation of the model was performed by comparing the predictions generated by the model with the manually annotated masks which were approved by a trained physician. 

\begin{table}[!ht]
\scriptsize
\centering
\caption{Prediction probabilities for non-bleeding and bleeding class for a few of the sample validation images.}
\begin{tabular}{l|l}
\hline
Bleeding images     & Prediction probability                        \\ \hline
Image 1             & \textit{{[}0.0000000e+00, 1.0000000e+00{]}}   \\ \cline{2-2} 
Image 2             & \textit{{[}0.0000000e+00, 1.0000000e+00{]}}   \\ \cline{2-2} 
Image 3             & \textit{{[}0.0000000e+00, 1.0000000e+00{]}}   \\ \cline{2-2} 
Image 4             & \textit{{[}0.0000000e+00, 1.0000000e+00{]}}   \\ \hline
Non-bleeding images & Prediction probability                        \\ \hline
Image 5             & \textit{{[}1.00000000e+00, 4.60899956e-31{]}} \\ \cline{2-2} 
Image 6             & \textit{{[}1.00000000e+00, 3.78204294e-24{]}} \\ \cline{2-2} 
Image 7             & \textit{{[}1.00000000e+00, 2.82393328e-32{]}} \\ \cline{2-2} 
Image 8             & \textit{{[}1.00000000e+00, 4.27774860e-29{]}} \\ \hline
\end{tabular}
\label{tab:prob}
\end{table}

We proceeded with the images classified as non-bleeding or bleeding images from the validation data set. On top of that we implemented three different interpretable Explainable AI algorithms and by that provided three different explanations for the images. We then conduct a user study to see how explanation generated with LIME and SHAP affects human decision-making.

\subsection{Explainability}
We decided to use three different explainable methods Local Interpretable Model Agnostic Explanations (LIME), Shapley Additive exPlanations (SHAP) and Contextual Importance and Utility (CIU). We implemented LIME and SHAP explainable method on Triton, high-performance computing cluster provided by Aalto University, using Python language whereas CIU explanations were generated using RStudio Version 1.2.1335. Apart from the use of explainable methods we also included a setting without explanations. There is no kind of explanation given for the displayed images in this setting.
For our empirical evaluation, we use the black-box XAI as a baseline. 

Figures below are visualising explanations of explainable methods for particular image. Explanations are outlining the key features (bleeding or non-bleeding areas) on the images, generated with the use of interpretable algorithms LIME, SHAP and CIU. With the confirmation from a professional physician and annotated masks manually made by trained physicians we made sure that the displayed explanations provided by algorithms are indeed at least partly marking the correct areas and can be regarded as understandable.

Explanations are given in the form of yellow outlined boundaries around the key features of the images that influenced the black box model's decision
\subsubsection{LIME explanations}
\begin{figure*}[!ht]
\centering
\begin{minipage}{.80\textwidth}
  \centering
  \includegraphics[width=1.0\linewidth]{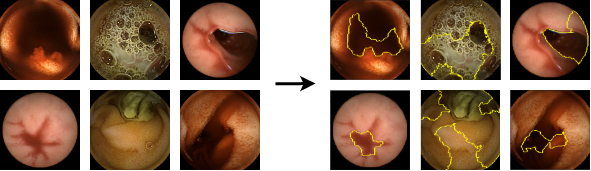}
  \captionof{figure}{LIME explanations.}
  \label{fig:lime_exp}
\end{minipage}%
\end{figure*}

LIME explanations were generated  with setting features; num\_samples (size of the neighborhood to learn the linear model) to value of 2500 and num\_features (maximum number of features present in explanation; number of superpixels to include in explanation) value of 10. LIME has been tested for all the images of the validation data set, both bleeding and non-bleeding. Figure \ref{fig:lime_exp} depicts some of the explanations provided by LIME. Explanations are given in the form of yellow outlined boundaries around the key features of the images that influenced the black box model's decision. In the case of the bleeding image LIME explanation is marking the area which positively contributes to the bleeding class and in the non-bleeding image LIME explanation is marking the area contributing to the non-bleeding class. 

\subsubsection{SHAP explanations}

\begin{figure*}[!ht]
\centering
\begin{minipage}{.80\textwidth}
  \centering
  \includegraphics[width=1.00\linewidth]{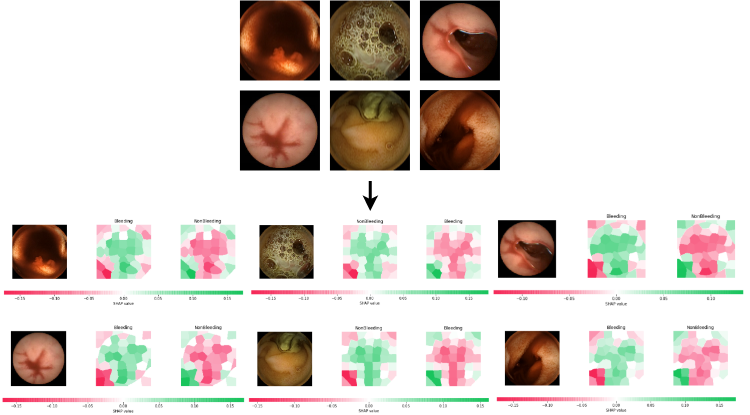}
  \captionof{figure}{SHAP explanations.}
  \label{fig:shap_exp}
\end{minipage}%
\end{figure*}

SHAP has as well been tested for all the images of the validation data set, both bleeding and non-bleeding, and the explanations provided by SHAP are depicted in Figure \ref{fig:shap_exp}. We applied the model agnostic Kernel SHAP method on a super-pixel segmented image to explain the convolutional neural network's image predictions. SHAP explanations were generated at the num\_samples (size of the neighborhood to learn the linear model) value of 2500. SHAP explanations are in each example of the image depicting contributions to both bleeding and non-bleeding class. Green color marking the important features of the images is representing a support for the class (bleeding, non-bleeding) and the red color is representing contradiction to the class (bleeding, non-bleeding).

\subsubsection{CIU explanations}

\begin{figure*}[!ht]
\centering
\begin{minipage}{.80\textwidth}
  \centering
  \includegraphics[width=1.0\linewidth]{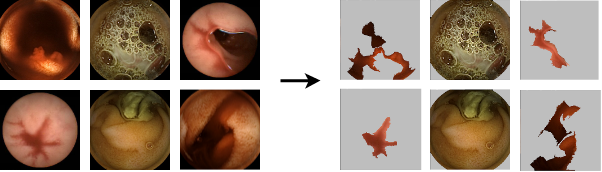}
  \captionof{figure}{CIU explanations.}
  \label{fig:ciu_exp}
\end{minipage}%
\end{figure*}

The CIU explanations were generated at the threshold value of 0.01 and 50 numbers of superpixels. CIU has also been tested for all the images of the validation data set, both bleeding and non-bleeding, and the CIU explanations are shown in Figure \ref{fig:ciu_exp}. CIU explanations are similarly to LIME marking the important area on the image which contributes to the given class, either the bleeding or the non-bleeding class. 

\section{Human evaluation user study}
To investigate the impact of the explainable machine learning methods on human decision-making we conducted three user studies based on the three explainable methods LIME, SHAP and CIU which provided explanations of the images from the medical image data set. Our aim was to see how well users understood the explainable decision-making support and to compare the utility of various explainable methods.  We also intended to analyze the satisfaction of users with explanations and figure out if users can recognize the effectiveness of the explanations.

\subsection{Data collection}
In order to generalize the use of explainable algorithms we decided to invite the normal users instead of experienced medical professionals. The reason for that was that explanations should be made as simple as possible in order to make them easily understandable even for the non-medical people. The users, in this case normal users completed a series of tests in the web-based survey and stated their experience and understanding of the presented explanations.

\begin{table}[!ht]
\scriptsize
\centering
\caption{Demographics of study participants from LIME, SHAP and CIU user study (all with the included noEXP testing).}
\begin{tabular}{l|r|rrr|rrr|rr|rr|l}
\hline
Methods                                                     & \multicolumn{1}{l|}{Total} & \multicolumn{1}{l}{Gender}           & \multicolumn{1}{l}{\textit{}}      & \multicolumn{1}{l|}{\textit{}}    & \multicolumn{1}{l}{\begin{tabular}[c]{@{}l@{}}Highest \\ degree\end{tabular}}    & \multicolumn{1}{l}{}                                                            & \multicolumn{1}{l|}{}                                                             & \multicolumn{1}{l}{\begin{tabular}[c]{@{}l@{}}STEM \\ background\end{tabular}} & \multicolumn{1}{l|}{}   & \multicolumn{1}{l}{\begin{tabular}[c]{@{}l@{}}XAI \\ understanding\end{tabular}} & \multicolumn{1}{l|}{}   & \begin{tabular}[c]{@{}l@{}}Age\\ (years)\end{tabular}                                                                                \\ \cline{3-12}
                                                            & \multicolumn{1}{l|}{}      & \multicolumn{1}{l|}{\textit{Female}} & \multicolumn{1}{l|}{\textit{Male}} & \multicolumn{1}{l|}{\textit{OTH}} & \multicolumn{1}{l|}{\begin{tabular}[c]{@{}l@{}}Ph.D \\ (or higher)\end{tabular}} & \multicolumn{1}{l|}{\begin{tabular}[c]{@{}l@{}}Master's \\ degree\end{tabular}} & \multicolumn{1}{l|}{\begin{tabular}[c]{@{}l@{}}Bachelor's \\ degree\end{tabular}} & \multicolumn{1}{l|}{Yes}                                                       & \multicolumn{1}{l|}{No} & \multicolumn{1}{l|}{Yes}                                                         & \multicolumn{1}{l|}{No} &                                                                                                                                      \\ \hline
\begin{tabular}[c]{@{}l@{}}LIME \\ (and noEXP)\end{tabular} & \textit{20}                & \multicolumn{1}{r|}{6}               & \multicolumn{1}{r|}{14}            & 0                                 & \multicolumn{1}{r|}{\textit{3}}                                                  & \multicolumn{1}{r|}{\textit{12}}                                                & \textit{5}                                                                        & \multicolumn{1}{r|}{\textit{19}}                                               & \textit{1}              & \multicolumn{1}{r|}{\textit{12}}                                                 & \textit{8}              & \textit{\begin{tabular}[c]{@{}l@{}}22, 23, 24, \\ 25(2), 26, \\ 27(4), 28, \\ 29, 30(2), \\ 31(3), 32, \\ 33, 34\end{tabular}}       \\ \hline
\begin{tabular}[c]{@{}l@{}}SHAP \\ (and noEXP)\end{tabular} & \textit{20}                & \multicolumn{1}{r|}{7}               & \multicolumn{1}{r|}{13}            & 0                                 & \multicolumn{1}{r|}{\textit{6}}                                                  & \multicolumn{1}{r|}{\textit{12}}                                                & \textit{2}                                                                        & \multicolumn{1}{r|}{\textit{18}}                                               & \textit{2}              & \multicolumn{1}{r|}{\textit{8}}                                                  & \textit{12}             & \textit{\begin{tabular}[c]{@{}l@{}}22(2), 23, \\ 24, 25, 26, \\ 27(4), 28, \\ 29(2), 30, \\ 31, 33, 36, \\ 38, 39, 42,\end{tabular}} \\ \hline
\begin{tabular}[c]{@{}l@{}}CIU \\ (and noEXP)\end{tabular}  & \textit{20}                & \multicolumn{1}{r|}{7}               & \multicolumn{1}{r|}{13}            & 0                                 & \multicolumn{1}{r|}{\textit{5}}                                                  & \multicolumn{1}{r|}{\textit{9}}                                                 & \textit{6}                                                                        & \multicolumn{1}{r|}{\textit{17}}                                               & \textit{3}              & \multicolumn{1}{r|}{\textit{9}}                                                  & \textit{11}             & \textit{\begin{tabular}[c]{@{}l@{}}21, 25(2), \\ 26(2), 27(3), \\ 28(3), 29(3), \\ 30(2), 32(2), \\ 34, 35\end{tabular}}             \\ \hline
\end{tabular}
\label{tab:demo}
\end{table}

For this user-centred studies, we gathered the participants from the University’s academic environment meaning that most of the participants have at least a Bachelor’s university degree. We collected data from a total of 60 (n=60) users, 20 users in each of the three test groups. 20 users performed testing with no explanation and LIME explanation support, 20 users performed testing with no explanation and SHAP explanation support, and 20 users performed testing with no explanation and CIU explanation support. Table \ref{tab:demo} is showing the demographics of the study participants. The users predominantly have master's education and STEM (science and technology) background, are in their twenties or thirties and are predominantly males. Approximately half of the participants have heard of the XAI prior to participating in the user study.

\begin{figure*}[!ht]
\centering
\begin{minipage}{0.50\textwidth}
  \centering
  \includegraphics[width=0.8\linewidth]{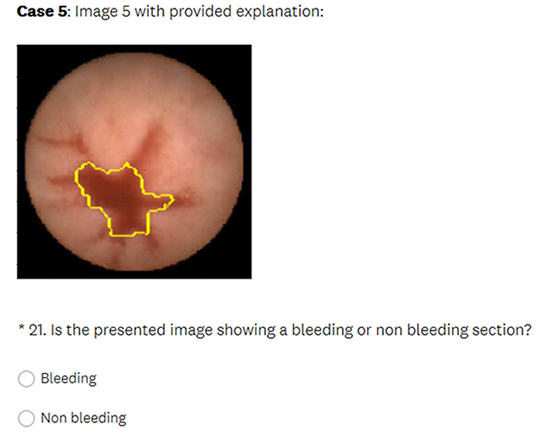}
  \captionof{figure}{LIME explanations inside the user study.}
  \label{fig:lime_study}
\end{minipage}%
\begin{minipage}{0.50\textwidth}
  \centering
  \includegraphics[width=0.8\linewidth]{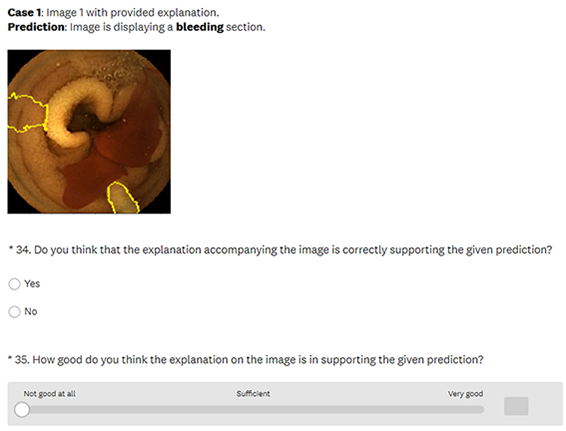}
  \captionof{figure}{Incorrect LIME explanations inside the user study.}
  \label{fig:lime_study2}
\end{minipage}
\end{figure*}

\begin{figure*}[!ht]
\centering
\begin{minipage}{0.50\textwidth}
  \centering
  \includegraphics[width=0.8\linewidth]{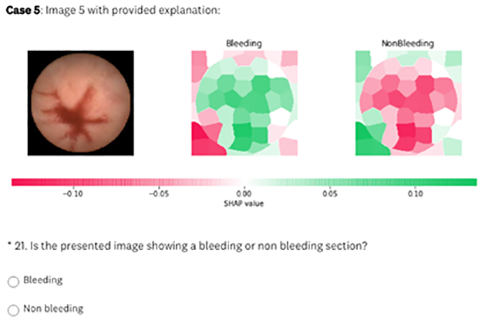}
  \captionof{figure}{SHAP explanations inside the user study.}
  \label{fig:shap_study}
\end{minipage}%
\begin{minipage}{0.50\textwidth}
  \centering
  \includegraphics[width=0.8\linewidth]{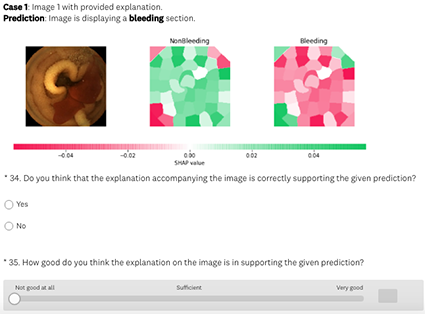}
  \captionof{figure}{Incorrect SHAP explanations inside the user study.}
  \label{fig:shap_study2}
\end{minipage}
\end{figure*}

\begin{figure*}[!ht]
\centering
\begin{minipage}{0.50\textwidth}
  \centering
  \includegraphics[width=0.8\linewidth]{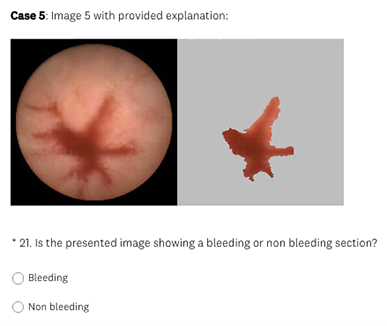}
  \captionof{figure}{CIU explanations inside the user study.}
  \label{fig:ciu_study}
\end{minipage}%
\begin{minipage}{0.50\textwidth}
  \centering
  \includegraphics[width=0.8\linewidth]{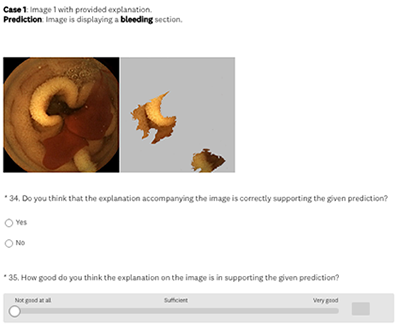}
  \captionof{figure}{Incorrect CIU explanations inside the user study.}
  \label{fig:ciu_study2}
\end{minipage}
\end{figure*}

\subsection{Study description and design of the user study}
\begin{figure*}[!ht]
\centering
\begin{minipage}{.80\textwidth}
  \centering
  \includegraphics[width=1.0\linewidth]{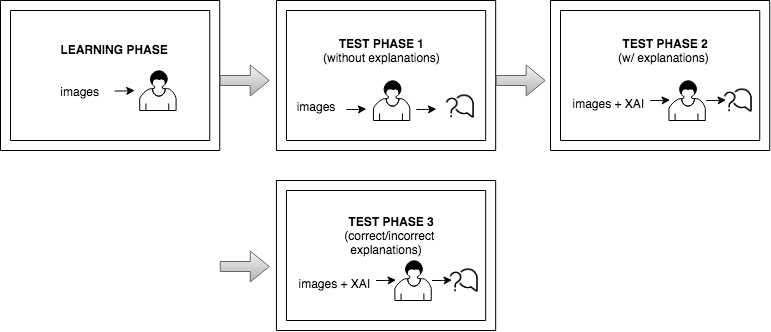}
  \captionof{figure}{User study design.}
  \label{fig:study_design}
\end{minipage}%
\end{figure*}

Our participants were distributed into three different groups. First group was presented with noXAI and LIME explanations, the second one with noXAI and SHAP explanations and the third group with noXAI and CIU explanations. The user studies designed for each of the three XAI methods are depicted in Figures \ref{fig:lime_study}, \ref{fig:shap_study} and \ref{fig:ciu_study}. The basic layout design for the user study is depicted in figure \ref{fig:study_design}. The figures \ref{fig:lime_study2}, \ref{fig:shap_study2} and \ref{fig:ciu_study2} are depicting the third part of the test phase where also the incorrect explanations are presented.

\medskip \textbf{Hypotheses.} The aim of this research study was to assess the following hypotheses: 
\begin{enumerate}
    \item \textbf{$H_{a}$:} Participants in the first group will perform better when given help of the LIME explanations.

\item \textbf{$H_{b}$:} Participants in the second group will perform better when given help of the SHAP explanations.

\item\textbf{$H_{c}$:} Participants in the third group will perform better when given help of the CIU explanations.

\item \textbf{$H_{d}$:} Participants given CIU explanation will perform better (make more correct decisions) than participants given LIME explanation.

\item \textbf{$H_{e}$:}  Participants given CIU explanation will perform (make more correct decisions) better than participants given SHAP explanation.

\item \textbf{$H_{f}$:} Participants given CIU explanations will understand  the explanations (distinguish correct explanations from incorrect) better than participants given LIME explanation.

\item \textbf{$H_{g}$:} Participants given CIU explanations will understand the explanations (distinguish correct explanations from incorrect) better than participants given SHAP explanation.
\end{enumerate}

We are testing whether we can reject the null hypotheses (Ha0, Hb0, Hc0, Hd0, He0, Hf0, Hg0) being the negations of our seven hypotheses. First three hypotheses focus on whether explanations are supporting humans in making the correct decisions in comparison with having no-explanation, and hypotheses from fourth to seventh focus on predicting the differences between the three different explanation supports. Fourth and fifth hypotheses are focused on differences in decision making between users having different explanation support. The last two hypotheses' aim was to evaluate whether human users are able to detect the bias by the provided explanations introduced in the five out of twelve test cases in the last part of the test phase and how the ability to recognize correct or incorrect explanations differs among the three user groups.

\begin{enumerate}
\item In the first stage we presented the used medical images to the study participants and explained them the essential information for completing the study together with the form of written instructions. 
\item After the users were familiarized with the required instructions the user study was carried out under the supervision of one of the researchers who was responsible for the control of study process. 
\item The user study started with the learning part where user was presented with a few test medical images with provided model's output in order for the user to learn to distinguish the bleeding from normal (non-bleeding) image. 
\item In the test phase the user was first presented with the medical images different from the ones used in the learning section and had to make a decision whether the images shown are displaying the bleeding or normal section. And in the next phase the user was presented with the same medical images as in the previous phase but this time together with the visual explanations provided by explainable method without the precise decision offered by a black-box model. The explanation is presented in such a way that the important features on the initial image from our data set, in this case normal or bleeding areas if any, are highlighted or isolated.  The diagnosis was set to be made by the user in both test phases. The user needed to decide if there is indeed any bleeding present and if the image shows severe condition or if there is no bleeding present. By that we could verify whether proposed explainable methods indeed enhanced number of correct decisions by the human. In the end of the second test phase the user had to rate the explanations by marking how satisfied they are with the explanations provided on a Likert scale from 0 to 5 (0 expressing the lowest satisfaction, and 5 the highest satisfaction with the explanations).
\item In the last part of the test phase users had to indicate if they thought that the presented explanation was correct or not for each presented image. By that we got information about users’ ability to understand the explanations by recognizing the cases where explanation by the explainable method was correct or incorrect.
\item The process was iterated for 4 different cases of data in the learning stage, 16 cases of data in the second and 12 in the third stage. Cases in the learning phase differed from the cases in the test phase. The presented images from first and second test phase also differed from the images in the third part of the test phase. In the test phase users were not allowed to reference the learning data.
\item After all the four rounds of the survey were completed, the users were also asked to complete the evaluation questionnaire through which we got their demographics information and information about their understanding of the explanation support. Because these questions could influence the users ’ perception of the procedure, they were only posed after the research evaluation was completed and could not be accessed by the user beforehand. 
\end{enumerate}

Throughout the trial, all participants in each user group obtained the same data points.
This architecture accounts for any data discrepancies between conditions and users.
We also matched the data across the no-explanation and explanation phases to control for the impact of specific data points on user accuracy. In order to isolate the impact of the explanation support we measured the users' initial accuracy prior to measuring the accuracy when they were given the help of the explanations.

\section{Analyses of the results}

\subsection {Performance of LIME, SHAP and CIU}
When generating explanations with the three different explainable methods we also set to compare their time needed for generating explanations and their overall performance. LIME needed around 11 seconds per image (11.4 seconds) and around 5 minutes and 20 seconds for 28 images with num\_samples=2500 and num\_features=10. For generating explanations on all the validation images (354 images) the timing needed is 1 hour and 45 minutes. SHAP needed around 10 seconds per image (9.8 seconds), and around 4 minutes, 35 seconds for 28 images with num\_samples=3000. This would be around 1 hour 30 minutes for all validation images. In comparison to SHAP and LIME, the running time of CIU explainable method  was about 8.5 seconds per image and the total time for producing explanations for 28 selected images from our data set was 4 minutes. For all validation images the timing would be around 1h and 18 minutes. 

\begin{table}[!ht]
\scriptsize
\centering
\caption{LIME, SHAP and CIU time comparison; (Note. LIME and SHAP were run using Python and CIU was run uging R).}
\begin{tabular}{l|l|l|l|l}
\hline
                                                             &            & LIME                & SHAP                    & CIU                \\ \hline
\begin{tabular}[c]{@{}l@{}}Timing \\ comparison\end{tabular} & 1 image  & \textit{11.40 sec.}    & \textit{9.80 sec.}      & \textit{8.50 sec.} \\ \cline{2-5} 
                                                             & 28 images  & \textit{5 min. 20 sec}    & \textit{4 min. 30 sec.} & \textit{4 min.}    \\ \cline{2-5} 
                                                             & 354 images & \textit{1h 45 min.} & \textit{1h 30 min}      & \textit{1h 18min}  \\ \hline
\end{tabular}
\label{}
\end{table}

Important to note is that experiments for LIME and SHAP were run using Triton, high-performance computing cluster provided by Aalto University whereas experiments for CIU were run using RStudio Version 1.2.1335 (R version 3.6.1) on a MacBook Pro, with 2,3 GHz 4-Core Intel Core i7 processor, 16 GB 1333 MHz DDR3 memory, and Intel HD Graphics 3000 512 mb graphics card.  When running LIME in R Studio, LIME took about 1 minute, 50 seconds for generating explanations on each image. SHAP has not been tested using R Studio.

\subsection{Quantitative analyses of explanations}

In order to explore the impact of explanations generated with the three explainable methods on the human decision-making we first analysed results for each group and compared the difference between users' performance in testing without explanation and testing where they had a help of explanations generated with one of the explainable methods. After that we compared three different user study groups: those who were given 1. LIME explanations, 2. SHAP explanations, and 3. CIU explanations.  We used IBM SPSS Statistics Version 23.0.0.0 to analyze the data, running hypotheses tests as well as exploratory statistics. By analyzing the data from the three user study settings we first examined the differences between means and medians of human decision-making (Table \ref{tab:measures}, Table \ref{tab:under}) and for each of the hypotheses tested the difference in human decision-making using Independent-Sample t-Test assuming unequal variances with significance level of $\alpha$ set to 0.05 (Table \ref{tab:hypo}).   

\subsubsection{Analyses of human decision-making from the three users groups with different explanation support methods}

\begin{table}[!ht]
\scriptsize
\centering
\caption{Mean and median values of users' decision making.}
\begin{tabular}{l|l|rr|rr|rr}
\hline
                   & Measures  & \multicolumn{1}{l}{\begin{tabular}[c]{@{}l@{}}LIME\\ user study\end{tabular}} & \multicolumn{1}{l|}{}                    & \multicolumn{1}{l}{\begin{tabular}[c]{@{}l@{}}SHAP\\ user study\end{tabular}} & \multicolumn{1}{l|}{}                    & \multicolumn{1}{l}{\begin{tabular}[c]{@{}l@{}}CIU\\ user study\end{tabular}} & \multicolumn{1}{l}{}                    \\ \cline{2-8} 
                   & \textit{} & \multicolumn{1}{l|}{\begin{tabular}[c]{@{}l@{}}With\\ explanation\end{tabular}}                                         & \multicolumn{1}{l|}{\begin{tabular}[c]{@{}l@{}}Without\\ explanation\end{tabular}} & \multicolumn{1}{l|}{\begin{tabular}[c]{@{}l@{}}With\\ explanation\end{tabular}}                                         & \multicolumn{1}{l|}{\begin{tabular}[c]{@{}l@{}}Without\\ explanation\end{tabular}} & \multicolumn{1}{l|}{\begin{tabular}[c]{@{}l@{}}With\\ explanation\end{tabular}}                                        & \multicolumn{1}{l}{\begin{tabular}[c]{@{}l@{}}Without\\ explanation\end{tabular}} \\ \hline
Correct decision   & Mean      & \multicolumn{1}{r|}{\textit{14.15}}                                           & \textit{13.95}                           & \multicolumn{1}{r|}{\textit{13.40}}                                           & \textit{14.05}                           & \multicolumn{1}{r|}{\textit{14.90}}                                          & \textit{14.30}                          \\ \cline{2-8} 
                   & Median    & \multicolumn{1}{r|}{\textit{14.50}}                                           & \textit{15.00}                           & \multicolumn{1}{r|}{\textit{15.00}}                                           & \textit{14.00}                           & \multicolumn{1}{r|}{\textit{16.00}}                                          & \textit{14.00}                          \\ \hline
Incorrect decision & Mean      & \multicolumn{1}{r|}{\textit{1.80}}                                            & \textit{2.05}                            & \multicolumn{1}{r|}{\textit{2.60}}                                            & \textit{1.95}                            & \multicolumn{1}{r|}{\textit{1.10}}                                           & \textit{1.70}                           \\ \cline{2-8} 
                   & Median    & \multicolumn{1}{r|}{\textit{1.50}}                                            & \textit{1.00}                            & \multicolumn{1}{r|}{\textit{1.00}}                                            & \textit{2.00}                            & \multicolumn{1}{r|}{\textit{0.00}}                                           & \textit{2.00}                           \\ \hline
\end{tabular}
\label{tab:measures}
\end{table}

\begin{table}[!ht]
\scriptsize
\centering
\caption{Mean and median values of users' ability to recognize correct and incorrect explanations.}
\begin{tabular}{l|l|r|r|r}
\hline
                                                                      & Measure & \multicolumn{1}{l|}{LIME} & \multicolumn{1}{l|}{SHAP} & \multicolumn{1}{l}{CIU} \\ \hline
\begin{tabular}[c]{@{}l@{}}Recognition of correct and \\ incorrect explanations
\end{tabular} & Mean    & \textit{8.85}             & \textit{8.65}             & \textit{10.25}          \\ \cline{2-5} 
                                                                      & Median  & \textit{9.50}             & \textit{9.50}             & \textit{11.00}          \\ \hline
\end{tabular}
\label{tab:under}
\end{table}

\begin{table}[!ht]
\scriptsize
\centering
\caption{Hypotheses analyses; (Note. *p$<$.05, **p$<$.01, ***p$<$.001).}
\begin{tabular}{l|l|l|r|r}
\hline
  & t-test        & Hypothesis & \multicolumn{1}{l|}{p-value(one-tailed)} & \multicolumn{1}{l}{p-value(two-tailed)} \\ \hline
1 & (LIME, noEXP) & Ha0        & \textit{0.334}                           & \textit{0.738}                          \\ \hline
2 & (SHAP, noEXP) & Hb0        & \textit{0.232}                           & \textit{0.464}                          \\ \hline
3 & (CIU, noEXP)  & Hc0        & \textit{0.079}                           & \textit{0.158}                          \\ \hline
4 & (CIU, LIME)   & Hd0        & \textit{0.059}                           & \textit{0.120}                          \\ \hline
5 & (CIU, SHAP)   & He0        & \textit{0.036*}                          & \textit{0.073}                          \\ \hline
6 & (CIU, LIME)   & Hf0        & \textit{0.009**}                         & \textit{0.018*}                         \\ \hline
7 & (CIU, SHAP)   & Hg0        & \textit{0.037*}                          & \textit{0.073}                          \\ \hline
\end{tabular}
\label{tab:hypo}
\end{table}

\begin{table}[!ht]
\scriptsize
\centering
\caption{Comparison between LIME and SHAP; (Note. *p$<$.05, **p$<$.01, ***p$<$ .001).}
\begin{tabular}{l|l|l|r|r}
\hline
  & t-test       &                                                                              & \multicolumn{1}{l|}{p-value(one-tailed)} & \multicolumn{1}{l}{p-value(two-tailed)} \\ \hline
1 & (LIME, SHAP) & Users' decision making                                                           & \textit{0.185}                           & \textit{0.370}                          \\ \cline{2-5} 
2 & (LIME SHAP)  & \begin{tabular}[c]{@{}l@{}}Recognizing of correct and \\  incorrect explanations \end{tabular} & \textit{0.414}                           & \textit{0.827}                          \\ \hline
\end{tabular}
\label{tab:lime_shap}
\end{table}

Table \ref{tab:measures} is showing the mean and median of correct and incorrect decisions for each type of explanations as well as no-explanations setting for all three user studies. Aligned with our first five hypotheses there are notable differences in means regarding the differences in decision making of users between different explanation settings as well as between explanation versus settings without explanation. Table \ref{tab:under} is showing notable differences in means in relation to our sixth and seventh hypotheses regarding the differences in users' understanding of explanations by distinguishing incorrect explanations from correct ones. The hypotheses analysis of all seven hypotheses is shown in Table \ref{tab:hypo}. Additionally Table \ref{tab:lime_shap} is showing the difference between users given LIME and users given SHAP explanation support.

\textbf{LIME.} Results from the users having LIME explanation support show that in comparison to the setting without explanation participants in line with our first hypothesis performed slightly better in the test phase with the explanations provided. Although the difference was not statistically significant (\textit{p} = 0.738) in receipt to the testing without explanation users answered more questions correctly (Table \ref{tab:measures}, Table \ref{tab:hypo}), in the test phase with LIME explanation support, meaning that they recognized that the image was displaying a bleeding or non-bleeding sequence in higher frequency. Results from the the third test phase show that participants were on average also able to recognize when the displayed LIME explanation was being correct or being incorrect. The mean of the correct decisions was 8.85 out of 12 answers in total (Table \ref{tab:under}).

\textbf{SHAP.} Results for SHAP show that users answered more questions correctly in testing without explanation, meaning that they recognized that the image was displaying a bleeding or non-bleeding image in higher frequency although the difference was not statistically significant (\textit{p} = 0.464) in receipt to the testing with explanation (Table \ref{tab:measures}, Table \ref{tab:hypo}). On the other hand participants were on average able to recognize when the displayed SHAP explanation was being correct or being incorrect. The mean of the correct answers was 8.47 out of 12 answers in total (Table \ref{tab:under}).

\textbf{CIU.} Results for CIU in line with our third hypothesis show that users answered more questions correctly in the test phase with explanation support, meaning that they recognized that the image was displaying a bleeding or non-bleeding image in higher frequency, although the difference was not statistically significant  (\textit{p} = 0.158) in receipt to the testing without explanation (Table \ref{tab:measures}, Table \ref{tab:hypo}). On average participants were really good at recognizing when the displayed CIU explanation was being correct or incorrect. The mean of the correct answers was 10.25 out of 12 answers in total (Table \ref{tab:under}). 

\textbf{Comparison between LIME and SHAP.} We performed a between-group comparison in order to find out which explanation supporting explainable methods (LIME, SHAP or CIU) was responsible for better decision-making of users. When comparing performance between LIME and SHAP, users having SHAP explanations reported higher satisfaction with the explanations, with the difference being statistically significant (\textit{p} = 0.0135, Table \ref{tab:comp}, Table \ref{tab:t-test}). Similarly users having SHAP explanation reported higher understanding of explanations in comparison to users with provided LIME explanations although the difference was not statistically significant (\textit{p} = 0.999, \ref{tab:comp}, Table \ref{tab:t-test}), Table \ref{tab:t-test}). However users given LIME explanation support answered more questions correctly in comparison to users having SHAP support although the difference is not statistically significant (\textit{p} = 0.370, Table \ref{tab:measures}, Table \ref{tab:lime_shap}). Users given SHAP explanations also spent significantly more time to complete the study in comparison to users having LIME explanation support (\textit{p} = 0.011, Table \ref{tab:comp}, Table \ref{tab:t-test}) which can indicate that SHAP explanations required more in depth concentration and were harder to interpret. When comparing if users were able to distinguish incorrect explanations from correct ones, the users given LIME explanation support performed better when compared to the users having SHAP explanations although the difference is not statistically significant (\textit{p} = 0.827, Table \ref{tab:lime_shap}).  

\textbf{Comparison between CIU, LIME and SHAP.} In comparison with users given LIME and those given SHAP explanations support, users given CIU explanation support answered more questions correctly although not significantly (\textit{p} = 0.120; \textit{p} = 0.073, Table \ref{tab:measures}, Table \ref{tab:hypo}). In terms of time for completing the study users given CIU explanation support spent significantly less time in comparison to users given SHAP explanations (\textit{p} = 0.016, Table \ref{tab:comp}, Table \ref{tab:t-test}) and more time in comparison to users given LIME explanations (\textit{p} = 0.633, Table \ref{tab:comp}, Table \ref{tab:t-test}) but in this case the time difference was not statistically significant. When comparing if users were able to distinguish incorrect explanations from correct ones, the users given CIU explanation support showed higher understanding of explanations in comparison to both users given LIME explanations (\textit{p} = 0.120, Table, \ref{tab:under}, Table \ref{tab:hypo}) and to users given SHAP explanation support (\textit{p} = 0.073, Table, \ref{tab:under}, Table \ref{tab:hypo}) although the difference is not statistically significant. Apart from that participants given CIU explanations answered more questions correctly when having a help of explanations than when having no-explanation support, although the difference is again not statistically significant (\textit{p} = 0.158, Table \ref{tab:measures}, Table \ref{tab:hypo}). Users also reported significantly higher satisfaction with CIU explanations compared to users having LIME explanations (\textit{p} = 0.000, Table \ref{tab:comp}, Table \ref{tab:t-test}) and higher but statistically not significant satisfaction compared to users having SHAP explanations (\textit{p} = 0.144, Table \ref{tab:comp}, Table \ref{tab:t-test}). Similarly users having CIU explanation reported higher understanding of explanations in comparison to users with provided LIME explanations (\textit{p} = 0.389, Table \ref{tab:t-test}) as well as higher than users having SHAP explanations (\textit{p} = 0.389, Table \ref{tab:t-test}) although the difference is not statistically significant. 


\begin{table}[!ht]
\scriptsize
\centering
\caption{Mean and median values of satisfaction, understanding and time spent.}
\begin{tabular}{l|l|r|r|r}
\hline
                                                                &         & \multicolumn{1}{l|}{LIME} & \multicolumn{1}{l|}{\textit{SHAP}} & \multicolumn{1}{l}{\textit{CIU}} \\ \hline
Satisfaction                                                    & Mean    & \textit{2}                & \textit{3.20}                       & \textit{3.75}                    \\ \cline{2-5} 
                                                                & Meadian & \textit{2}                & \textit{3}                         & \textit{4}                       \\ \hline
\begin{tabular}[c]{@{}l@{}}Time spent \\ (minutes)\end{tabular} & Mean    & \textit{15.57}            & \textit{23.18}                     & \textit{16.30}                    \\ \cline{2-5} 
                                                                & Median  & \textit{14.83}            & \textit{21.18}                     & \textit{15.73}                   \\ \hline
Understanding                                                   & Yes     & \textit{16}               & \textit{16}                        & \textit{18}                      \\ \cline{2-5} 
                                                                & No      & \textit{4}                & \textit{4}                         & \textit{2}                       \\ \hline
\end{tabular}
\label{tab:comp}
\end{table}

\begin{table}[!ht]
\scriptsize
\centering
\caption{T-tests: Satisfaction, understanding and time spent; (Note. *p$<$.05, **p$<$.01, ***p$<$.001).}
\begin{tabular}{l|l|l|l}
\hline
                                                                & t-test       & p-value(one-tailed) & p-value(two-tailed) \\ \hline
Satisfaction                                                    & (LIME, SHAP) & \textit{0.007**}    & \textit{0.0135*}     \\ \cline{2-4} 
                                                                & (LIME, CIU)  & \textit{0.000***}   & \textit{0.000***}   \\ \cline{2-4} 
                                                                & (SHAP, CIU)  & \textit{0.072}      & \textit{0.144}      \\ \hline
Understanding                                                   & (LIME, SHAP) & \textit{0.499}      & \textit{0.999}      \\ \cline{2-4} 
                                                                & (LIME, CIU)  & \textit{0.195}      & \textit{0.389}      \\ \cline{2-4} 
                                                                & (SHAP, CIU)  & \textit{0.195}      & \textit{0.389}      \\ \hline
\begin{tabular}[c]{@{}l@{}}Time spent \\ (minutes)\end{tabular} & (LIME, SHAP) & \textit{0.005**}    & \textit{0.011**}    \\ \cline{2-4} 
                                                                & (LIME, CIU)  & \textit{0.317}      & \textit{0.633}      \\ \cline{2-4} 
                                                                & (SHAP, CIU)  & \textit{0.008**}    & \textit{0.0166**}   \\ \hline
\end{tabular}
\label{tab:t-test}
\end{table}

\subsubsection{Correlation analyses}

We also performed correlation analyses between users’ performance (count of the correct decisions) and different demographic variables (age, gender, education level, STEM background, knowledge of XAI), as well as between their performance and the time spent for completing the study and the users’ understanding and satisfaction with the explanations. The correlation was calculated using Spearman's rank correlation coefficient. Spearman's correlation was chosen because it captures the monotonic relationship between the variables instead of only a linear relationship and also works well with categorical variables like gender. Spearman's correlation coefficient values for all conditions with LIME, SHAP and CIU explanation support together witht he p-values are shown in Table \ref{tab:corr_xai}. Table \ref{tab:corr_noxai} is showing correlation analyses for the settings without explanation. 

\begin{table}[!ht]
\scriptsize
\centering
\caption{Correlation between demographic variables and decision making in XAI settings; (Note. *p$<$.05, **p$<$.01, ***p$<$ .001).}
\begin{tabular}{l|l|r|r|r}
\hline
Demographics    &                     & \multicolumn{1}{l|}{LIME} & \multicolumn{1}{l|}{SHAP} & \multicolumn{1}{l}{CIU} \\ \hline
Age             & correlation         & \textit{-0.451}           & \textit{0.229}            & \textit{-0.413}                   \\ \cline{2-5} 
                & p value(two-tailed) & \textit{0.046*}           & \textit{0.332}            & \textit{0.071}                    \\ \hline
Gender          & correlation         & \textit{0.349}            & \textit{0.272}            & \textit{-0.02}                    \\ \cline{2-5} 
                & p value(two-tailed) & \textit{0.132}            & \textit{0.246}            & \textit{0.934}                    \\ \hline
Education       & correlation         & \textit{-0.073}           & \textit{0.210}            & \textit{-0.321}                   \\ \cline{2-5} 
                & p value(two-tailed) & \textit{0.760}            & \textit{0.374}            & \textit{0.167}                    \\ \hline
STEM background & correlation         & \textit{0.387}            & \textit{-0.309}           & \textit{-0.132}                   \\ \cline{2-5} 
                & p value(two-tailed) & \textit{0.092}            & \textit{0.185}            & \textit{0.580}                    \\ \hline
XAI             & correlation         & \textit{-0.235}           & \textit{0.274}            & \textit{-0.435}                   \\ \cline{2-5} 
                & p value(two-tailed) & \textit{0.318}            & \textit{0.242}            & \textit{0.055}                    \\ \hline
Time spent      & correlation         & \textit{-0.271}           & \textit{0.480}            & \textit{-0.10}                    \\ \cline{2-5} 
                & p value(two-tailed) & \textit{0.248}            & \textit{0.840}            & \textit{0.674}                    \\ \hline
Satisfaction    & correlation         & \textit{-0.519}           & \textit{0.482}            & 0.396                             \\ \cline{2-5} 
                & p value(two-tailed) & \textit{0.019*}           & \textit{0.031*}           & 0.084                             \\ \hline
Understanding   & correlation         & \textit{-0.377}           & \textit{0.522}            & 0.188                             \\ \cline{2-5} 
                & p value(two-tailed) & \textit{0.101}            & \textit{0.018}            & 0.427                             \\ \hline
\end{tabular}
\label{tab:corr_xai}
\end{table}

\begin{table}[!ht]
\scriptsize
\centering
\caption{Correlation between demographic variables and decision making in noEXP settings; (Note. *p$<$.05, **p$<$.01, ***p$<$ .001).}
\begin{tabular}{l|l|r|r|r}
\hline
Demographics    &                     & \multicolumn{1}{l|}{noEXP(LIME)} & \multicolumn{1}{l|}{noEXP(SHAP)} & \multicolumn{1}{l}{noEXP(CIU)} \\ \hline
Age             & correlation         & \textit{-0.18}                   & \textit{-0.063}                  & \textit{-0.318}                          \\ \cline{2-5} 
                & p value(two-tailed) & \textit{0.447}                   & \textit{0.792}                   & \textit{0.171}                           \\ \hline
Gender          & correlation         & \textit{0.069}                   & \textit{-0.166}                  & \textit{0.313}                           \\ \cline{2-5} 
                & p value(two-tailed) & \textit{0.773}                   & \textit{0.484}                   & \textit{0.179}                           \\ \hline
Education       & correlation         & \textit{0.173}                   & \textit{-0.246}                  & \textit{-0.087}                          \\ \cline{2-5} 
                & p value(two-tailed) & \textit{0.465}                   & \textit{0.296}                   & \textit{0.715}                           \\ \hline
STEM background & correlation         & \textit{0.392}                   & \textit{0.217}                   & \textit{0.139}                           \\ \cline{2-5} 
                & p value(two-tailed) & \textit{0.087}                   & \textit{0.357}                   & \textit{0.558}                           \\ \hline
XAI             & correlation         & \textit{0}                       & \textit{-0.285}                  & \textit{-0.264}                          \\ \cline{2-5} 
                & p value(two-tailed) & \textit{1}                       & \textit{0.223}                   & \textit{0.261}                           \\ \hline
Time spent      & correlation         & \textit{-0.191}                  & \textit{-0.491}                  & \textit{-0.191}                          \\ \cline{2-5} 
                & p value(two-tailed) & \textit{0.419}                   & \textit{0.028*}                  & \textit{0.419}                           \\ \hline
\end{tabular}
\label{tab:corr_noxai}
\end{table}

\textbf{LIME.} In the case of LIME we found a significant correlation, between users' count of the correct decisions and their age (\textit{p} = 0.046, Table \ref{tab:corr_xai}) which may be indicating that the participants with a lower age achieved a higher number of correct decisions, and were more satisfied with the provided explanations (\textit{p} = 0.019, Table \ref{tab:corr_xai}). Interestingly the lower was the users' satisfaction the better the users were in making the decision, which may indicate that the users who recognized LIME support in decision also recognized that LIME could do better job in decision support. In the setting without explanation we did not find any significant correlation (Table \ref{tab:corr_noxai}).

\textbf{SHAP.} In the setting with SHAP explanation support we found correlation between users' count of the correct decisions and higher satisfaction with SHAP explanations (\textit{p} = 0.031, Table \ref{tab:corr_xai}) as well as with stating that they understood SHAP better (\textit{p} = 0.018, Table \ref{tab:corr_xai}). This may indicate that users in general were satisfied with the help of SHAP explanations and their appearance, however due to SHAP explanations being presented in a more complex way it appears that users had harder time understanding them. In the setting without explanation we found significant correlation between users' count of the correct decisions and less time needing for for completing the study (Table \ref{tab:corr_noxai}).

\textbf{CIU.} We did not find any significant correlation with the given CIU explanation support, as well as in the setting without the explanations provided (Table \ref{tab:corr_xai}, Table \ref{tab:corr_noxai}). 


\subsection{Qualitative analyses of explanations} Most of the participants with provided LIME, SHAP or CIU explanation, answered that they were able to understand the provided CIU explanations which as shown in the Table \ref{tab:comp} and Table \ref{tab:t-test} are having the highest rating. Users rating of the explanations show that they were the most satisfied with the explanations generated by CIU and less with those of SHAP or LIME. They were mostly satisfied with the provided explanations but also noted that the explanation could be more precise and should cover a bigger area of the image.
By analyzing the participants’ feedback statements from the evaluation questionnaire, we also found that:
\begin{enumerate}
\item Users want more precise identification of the important area on some of the presented images. 
\item In addition to visual explanations, users want supplementary text explanations. 
\item Users would like to have an option of interacting with the explainable method in order to get more in-depth information.
\end{enumerate}


\section{Discussion}
In the present study we observed notable differences when comparing human decision-making from the three groups of users with different explanation support. The observed differences are reflecting our initial assumptions stating that users having CIU explanation support will perform better than those having SHAP or LIME explanation support. Our results suggest that CIU was of bigger help to users in making the correct decisions, who were also more satisfied with its presentation in comparison to users having LIME and those having SHAP support. This indicates that explanations generated with CIU appeared more clear to the users and thus showed a greater support in decision making. 

Our results also support the last two hypothesis stating that participants with CIU will perform better in understanding the provided explanations and by that better distinguish between correct and incorrect explanation in comparison to participants having LIME or SHAP explanation support. Users with CIU explanation support were significantly better at recognizing incorrect explanations in comparison to those having LIME explanations and also to some extent better than those having SHAP explanation support. Additionally users given CIU explanation support spent significantly less time to complete the user study in comparison to users given SHAP explanations and more time in comparison to users given LIME explanations but in this case the time difference was not statistically significant. A possible explanation of this results is that participants with CIU explanation were able to better understand the provided explanations due to CIU explanations being depicted in lower complexity compared to those of SHAP and their more accurate representation of the significant area on the images compared to those of LIME. This assumptions can also concluded from the users' statements regarding their satisfaction and understanding of the explanation support. 

The present results also provide insight to the initial research questions concerning the use of the XAI methods in a form of a human decision-support and their contribution to the increased trust in AI-based Computer Vision systems in medical domain. The user studies and testing done with no explanation in comparison to setting with explanation support depicts that users are better at decision-making with the help of the explainable methods and are comfortable with the explanations provided. In two out of three user studies with each providing one of the explanations support (LIME, CIU) the participants performed better when the explanation support was provided. This was however not the case in the user study with provided SHAP explanations. Although users given SHAP explanations were relatively good at recognizing correct and incorrect explanations, the explanations by SHAP did not prove to increase the number of correct decisions in comparison to the no-explanation setting. In comparison to LIME and CIU user study the number of correct answers in both setting from SHAP user study is pretty low, however the setting without explanation proved to bring higher number of correct decisions of users, which can indicate that SHAP explanations were harder to understand and were confusing to the users. 

\subsection{Limitations}
The present paper provides evaluation of explainable methods (LIME, SHAP and CIU) support in human decision making.
The present study however has a set of limitations, with the most important ones listed below:
\begin{enumerate}
\item The current study's focus is limited to a single medical data set. The current use of explanation support focuses only on one set of medical images, which can further be tested on other more complex medical cases in need of a decision-support in various diagnosing.
\item It is important to expand the current scope of the studied data and apply the explanations to the actual real-life settings. By using the data in the sense of real world scenarios in actual real-life settings we may facilitate real-world applicability.
\item The scope of evaluation was limited to human evaluation testing, which requires basic tests with lay persons, due to time constraints: The study can be generalized to carry out application-based evaluations concerning real tasks performed by domain experts. In the case of the medical data, the most suitable users would be physicians working in diagnostics. 
\end{enumerate}

\subsection{Future work}

To overcome the limitations addressed in the previous sub-section, the following research directions may be considered in the future: 
\begin{enumerate}
\item With the future improvements we aim to generalize the explanations provided by explainable methods (LIME, SHAP and CIU) on different medical data sets and by that provide a wide decision-support for medical experts.
\item Currently for the human assessment we reduced the number of participants to 60 (20 for each case). Validating the results with a wider sample size would be rational. In addition, increasing the sample size could help produce more statistically meaningful hypotheses outcomes.
\item In order to provide a better evaluation and test the usability of the explanations, a user evaluation study with domain (medical) experts would be required. Furthermore the explanations could potentially be tested using application-based assessment, which would require domain experts doing activities related to the use of the explanations. 
\item Potentially we aim to expand out current work by dealing with the real-life case scenarios. It would be interesting to work with the real-world complexities in order to show that the explainable methods can help people make better decision in real-world situations.
\end{enumerate}

\section{Conclusion}
In line with our results, we present three potential explainable methods that can with future improvements in implementation be generalized on different medical data sets and can provide great decision-support for medical experts. From the viewpoint of users, our research offers a deep insight into the details of the explanation support that can be used as a constructive feedback for the potential implementation of machine learning explainable methods in the future. Our findings suggest that there are notable differences in human decision-making between various explanation support settings, with CIU showing the highest decision-support and performance. Additionally explanation support in comparison to the no-explanation provided, proved to increase the number of correct decisions made by the users in the two out of three user studies. The presented work thus gives developers more confidence to further develop and utilize the explainable methods which will instill more confidence and trust into users. Since in the present study the help of explanations was evaluated using human evaluation which involved lay humans in the future, we could aim to evaluate them using application-grounded evaluation which would involve domain experts, performing tasks specific to the use of the explanations.
By doing so, a clearer evaluation of the explanations could be obtained, and the explanations could be applied to other medical image processing situations with the future improvements and utilization. With the application of the explainable methods to the other medical data sets, as well as testing and improving them in the context of providing the decision support to medical professionals and automating diagnostic procedures may lead to solutions that are more broadly applicable.

\section*{Acknowledgements}
The research leading to this publication is supported by Helsinki Institute for Information Technology (grant 9160045), under the Finnish Center for Artificial Intelligence (FCAI) unit.

 \bibliographystyle{splncs04}
 \bibliography{ref}

\end{document}